\documentclass{article}

\usepackage{arxiv}

\usepackage[utf8]{inputenc} 
\usepackage[T1]{fontenc}    
\usepackage{hyperref}       
\usepackage{url}            
\usepackage{booktabs}       
\usepackage{amsfonts}       
\usepackage{nicefrac}       
\usepackage{microtype}      
\usepackage{lipsum}		
\usepackage{graphicx}
\usepackage{natbib}
\usepackage{doi}
\usepackage{amssymb}            
\usepackage{mathtools}          
\usepackage{mathrsfs}           
\mathtoolsset{showonlyrefs}     
\usepackage{graphicx}           
\usepackage{subcaption}         
\usepackage[space]{grffile}     
\usepackage{url}                
\usepackage{wrapfig}
\usepackage{algorithm2e}[rule]
\RestyleAlgo{ruled}
\SetKwComment{Comment}{/* }{ */}

\title{Multi-Agent Decision Transformers for Dynamic Dispatching in Material Handling Systems Leveraging Enterprise Big Data}

\author{
    Xian Yeow Lee  \\
    xian.lee@hal.hitachi.com \\
    Industrial A.I. Lab.\\
    Hitachi America Ltd.
    \And
    Haiyan Wang \\
    haiyan.wang@hal.hitachi.com \\
    Industrial A.I. Lab.\\
    Hitachi America Ltd.
    \And
    Daisuke Katsumata  \\
    daisuke.katsumata@hal.hitachi.com \\
    JR Automation Collaboration Project \\
    Hitachi America Ltd.
    \And
    Takaharu Matsui \\
    takaharu.matsui@hal.hitachi.com \\
    JR Automation Collaboration Project\\
    Hitachi America Ltd.
    \And
    Chetan Gupta  \\
    chetan.gupta@hal.hitachi.com \\
    Industrial A.I. Lab.\\
    Hitachi America Ltd.
    }

\date{}

\hypersetup{
pdftitle={A template for the arxiv style},
pdfsubject={q-bio.NC, q-bio.QM},
pdfauthor={David S.~Hippocampus, Elias D.~Striatum},
pdfkeywords={First keyword, Second keyword, More},
}

\begin{document}
\maketitle

\begin{abstract}
Dynamic dispatching rules that allocate resources to tasks in real-time play a critical role in ensuring efficient operations of many automated material handling systems across industries. Traditionally, the dispatching rules deployed are typically the result of manually crafted heuristics based on domain experts' knowledge. Generating these rules is time-consuming and often sub-optimal. As enterprises increasingly accumulate vast amounts of operational data, there is significant potential to leverage this big data to enhance the performance of automated systems. One promising approach is to use Decision Transformers, which can be trained on existing enterprise data to learn better dynamic dispatching rules for improving system throughput. In this work, we study the application of Decision Transformers as dynamic dispatching policies within an actual multi-agent material handling system and identify scenarios where enterprises can effectively leverage Decision Transformers on existing big data to gain business value. Our empirical results demonstrate that Decision Transformers can improve the material handling system's throughput by a considerable amount when the heuristic originally used in the enterprise data exhibits moderate performance and involves no randomness. When the original heuristic has strong performance, Decision Transformers can still improve the throughput but with a smaller improvement margin. However, when the original heuristics contain an element of randomness or when the performance of the dataset is below a certain threshold, Decision Transformers fail to outperform the original heuristic. These results highlight both the potential and limitations of Decision Transformers as dispatching policies for automated industrial material handling systems.
\end{abstract}

\section{Introduction}

Dynamic dispatching, the process of dispatching resources in real-time in response to system conditions, plays a critical role in ensuring smooth and efficient operations in many industrial applications. Subsequently, this translates to additional business value, such as cost savings and increased customer satisfaction. One area where the deployment of dynamic dispatching has a large impact is in material handling systems, where goods are typically transported between multiple points under the constraint of limited resources.

Traditionally, dynamic dispatching in these systems is often deployed using heuristic rules manually designed via a trial-and-error process or by a domain expert. Dynamic dispatching and scheduling are also applied in a wide variety of fields, not just material handling systems, and there have been multiple works that attempt to generate dispatching rules via domain knowledge and heuristics~\cite{dhurasevic2018survey, branke2015automated}. These approaches are time-consuming, and the availability of domain experts is usually not guaranteed. Another approach to deploying dynamic dispatching strategies is to employ optimization-based methods~\cite{qin2021dynamic, zhang2023survey}. Nevertheless, a major limitation of this category of approaches is that they can be time-consuming and computationally inefficient to compute a solution whenever a dispatch decision is required.

More recently, data-driven methods such as machine learning (ML) and reinforcement learning (RL)-based approaches has also been proposed~\cite{priore2014dynamic, ding2023survey}. While ML-based approaches focuses on imitating the performance of static datasets, RL-based dynamic dispatching policies explore a given environment during training to discover better policies~\cite{kang2019dynamic, jeong2021reinforcement, zeng2023deep, lee2024multi}. As such, RL-based approaches avoid the lengthy trial-and-error process of designing heuristics and circumvent the need to solve an optimization every time dispatching is needed.
However, the training process remains the main bottleneck in deploying RL-based policies. It is often not feasible to train RL policies on actual systems due to safety considerations~\cite{gu2022review} and due to the lengthy training duration required because of the large number of interactions typically necessary to train a good policy. Subsequently, most RL policies are trained on simulators, which provide a safe place for RL policies to explore and enable training beyond real-time speed. Nevertheless, developing a simulator is also often costly, and RL policies trained in simulator environments often suffer from the sim-to-real gap~\cite{zhao2020sim, salvato2021crossing}.

To address the complexity of training online RL-based algorithms, numerous works in the literature have advocated the idea of offline RL: that is, how to maximally extract an optimal policy given a static dataset~\cite{prudencio2023survey, levine2020offline}. With the proliferation of big data in industrial settings, vast amounts of historical operational data are often available, providing a rich resource for training more effective and robust policies and potentially leading to improved system performance. In the context of using RL for dynamic dispatching in material handling systems, this is a compelling paradigm, as a dataset could be collected based on historical real-time data using a sub-optimal dispatching policy, which could then be used to train a better RL-based policy. 

Unfortunately, offline-RL methods are known to be over-optimistic in terms of value estimation~\cite{kumar2020conservative} and high variance of reward/gradient estimation~\cite{levine2020offline}, which could result in destabilized policy training. In response, many efforts in the community have been directed at developing better offline-RL methods to mitigate these issues while still leveraging the benefits of historical data and RL-based approaches~\cite{levine2020offline, agarwal2020optimistic, prudencio2023survey}. 

Recently,~\cite{decision_transformer} proposed to reformulate the offline RL problem into a sequence modeling problem by leveraging the powerful modeling capabilities of a transformer architecture~\cite{vaswani2017attention}. They proposed Decision Transformers, which have been shown to be a strong alternative to existing offline RL algorithms without explicitly learning a value function~\cite{bhargava2023sequence}, and has been applied in multiple areas of work~\cite{yuan2024transformer}. As most existing work of Decision Transformers are mainly focused on benchmark problems, in this work, we are motivated to study if Decision Transformers can also be applied effectively to complex, real-world industrial problems due to their ability to leverage existing enterprise data, simplicity in terms of implementation and potential to be trained in a large scale, parallelized setting. Specifically, we investigate the feasibility of using Decision Transformers as a first step in developing data-driven dynamic dispatching strategies in a multi-agent setting to maximize the throughput of a material handling system. Our contributions are focused on answering the following questions: 
\begin{itemize} 
\item Can Decision Transformers train on existing enterprise operational data to discover more effective policies within a complex real-world material handling system where dispatching decisions are required at different points of the system asynchronously, i.e., given the multi-agent asynchronous setting, can independent Decision Transformers ``stitch” lower reward trajectories from sub-optimal heuristics to achieve higher rewards during testing?
\item Previous work by~\cite{paster2022you} has shown that Decision Transformers performs badly when trained on stochastic environments and we aim to study how are independent Decision Transformers deployed in multi-agent settings affected by environmental and data stochasticity? 
\item How do datasets generated from different heuristic qualities affect the final performance of the Decision Transformers? 
\end{itemize}

\section{Background}

\subsection{Material Dispatching Simulator}

\begin{figure}[h]
    \centering
    \includegraphics[width=0.7\columnwidth]{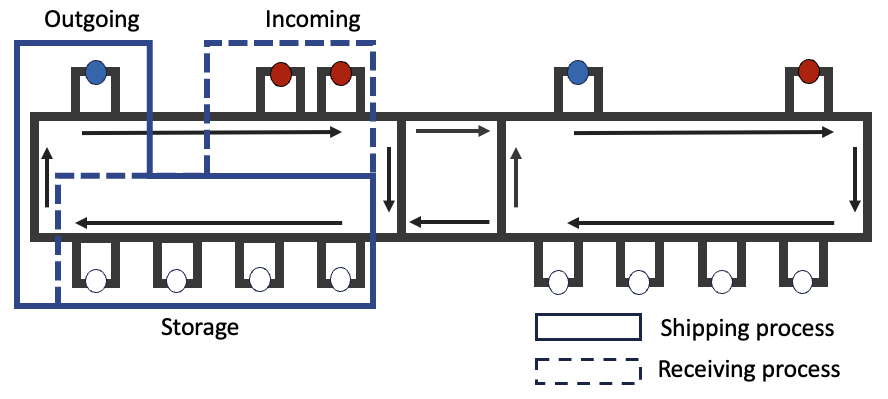}
    \caption{A simplified illustration of the material handling system we used in this case study.}
    \label{fig:system}
\end{figure}

In this section, we describe an instantiation of an actual material handling system used in this study. Specifically, we study a material handling system with a conveyor belt commonly used in warehouse facilities. To explain the system, a simplified version of the system layout is shown in Figure~\ref{fig:system}. In general, the material handling system consists of two processes: shipping and receiving; and three types of process points: incoming, storage and outgoing. In the receiving process, goods enter the material handling system via the incoming process points (red circles). These goods are placed on empty pallets on the conveyor belt and dynamically dispatched to one of the many possible storage points (white circles). Simultaneously, in the shipping process, goods are retrieved from the storage points, placed on empty pallets, and sent to the outgoing points (blue circles) to satisfy downstream tasks, such as fulfilling a customer order. A common Key Performance Index (KPI) for optimizing these systems is to maximize the throughput of the entire system; in this case, the combined throughput of goods entering the storage points and the goods being sent to the outgoing points, where throughput is the defined as the amount of material or items passing through a process over a specified period of time. In the shipping process, the storage points where goods are retrieved and the outgoing points where the goods are sent are specified by the downstream demand requests. However, in the receiving process, which storage point the goods should be dispatched to when it arrives at an incoming point should be carefully determined in real-time in order to optimize the KPI. The dispatching decisions are traditionally made by heuristics that are either crafted by domain experts or built through a trial-and-error process. 

These systems typically come with multiple constraints, making the dynamic dispatching decisions at incoming points complex. One of the main bottlenecks of optimizing the throughput of this system is that all the point processes share a common resource: the pallets circulating the system on the conveyor. Each point process also has a designated buffer, which only allows a certain number of incoming pallets to wait in a queue, and any additional pallets dispatched to the point process are circulated around the conveyor belt until the buffer has a vacant space. Additionally, there is an aspect of stochasticity present in the system in terms of the goods being retrieved from the storage that is dependent on the demand of the outgoing points. Together, these constraints give rise to numerous complexities, such as congestion at process points if too many pallets are dispatched to the same process points or idling process points when there are insufficient empty pallets for the process points to load and unload goods onto the pallets, thus negatively impacting the throughput of the system. Hence, deploying a good dynamic dispatching policy is imperative to maximizing the system's throughput. In this system, we consider dynamic dispatching policies that can be deployed at each of the incoming points, where the decision space of each policy consist of all the storage points.

\subsection{Decision Transformers}

Decision Transformers, introduced by Chen et al.~\cite{decision_transformer}, reformulate the offline reinforcement learning (RL) problem into a sequence modeling problem, leveraging the auto-regressive architecture of GPT-2~\cite{radford2019language}. The model takes as input a sequence, of length $k$, of past states \( \{s_{t-k}, \dots, s_t\} \), actions \( \{a_{t-k}, \dots, a_{t-1}\} \), and returns-to-go \( \{R_{t-k}, \dots, R_t\} \), where \( R_t = \sum_{t'=t}^T r_{t'} \), representing future cumulative rewards. At each time step \( t \), the input tuple \( \mathbf{x}_t = (s_t, a_{t-1}, R_t) \) is encoded into a sequence \( \mathbf{X}_t = \left[\mathbf{x}_0, \dots, \mathbf{x}_t\right] \), which is processed by a series of transformation to produce hidden states \( \mathbf{H}_t = \text{Transformer}(\mathbf{X}_t) \). The model predicts the next action \( \hat{a}_t \) by applying a linear transformation and softmax function to the final hidden state \( \mathbf{h}_t \), so that \( \hat{a}_t = \text{softmax}(W \mathbf{h}_t + b) \), where \( W \) and \( b \) are learned parameters. The model is trained using a supervised loss function, conventionally the mean squared error for continuous actions:
\[
\mathcal{L}_{\text{supervised}} = \mathbb{E}_{(s, a, R) \sim D} \left[ \sum_{t=0}^T \left\| a_t - \hat{a}_t \right\|^2 \right]
\]or the cross-entropy loss for discrete actions:
\[
\mathcal{L}_{\text{supervised}} = \mathbb{E}_{(s, a, R) \sim D} \left[ \sum_{t=0}^T - \log(\hat{p}_t[a_t]) \right]
\]
where \( a_t \) is the true action at time \( t \), \( \hat{a}_t \) is the predicted continuous action, \( \hat{p}_t \) is the predicted probability distribution for discrete actions, and \( \hat{p}_t[a_t] \) is the predicted probability for the true discrete action \( a_t \).

During deployment, it is conditioned on the initial state \( s_0 \) and desired cumulative reward \( R_0 \) to predict the first action, then autoregressively generates subsequent actions, each conditioned on the updated sequence of past states, actions, and remaining rewards, with \( R_{t+1} = R_t - r_t \). This method offers significant advantages over traditional RL methods, particularly in offline settings where the model learns from static datasets without exploration, and has demonstrated competitive performance on several benchmarks. For further details, we refer readers to the original paper~\cite{decision_transformer}.

\begin{figure*}[htbp]
    \centering
    \includegraphics[width=0.8\textwidth]{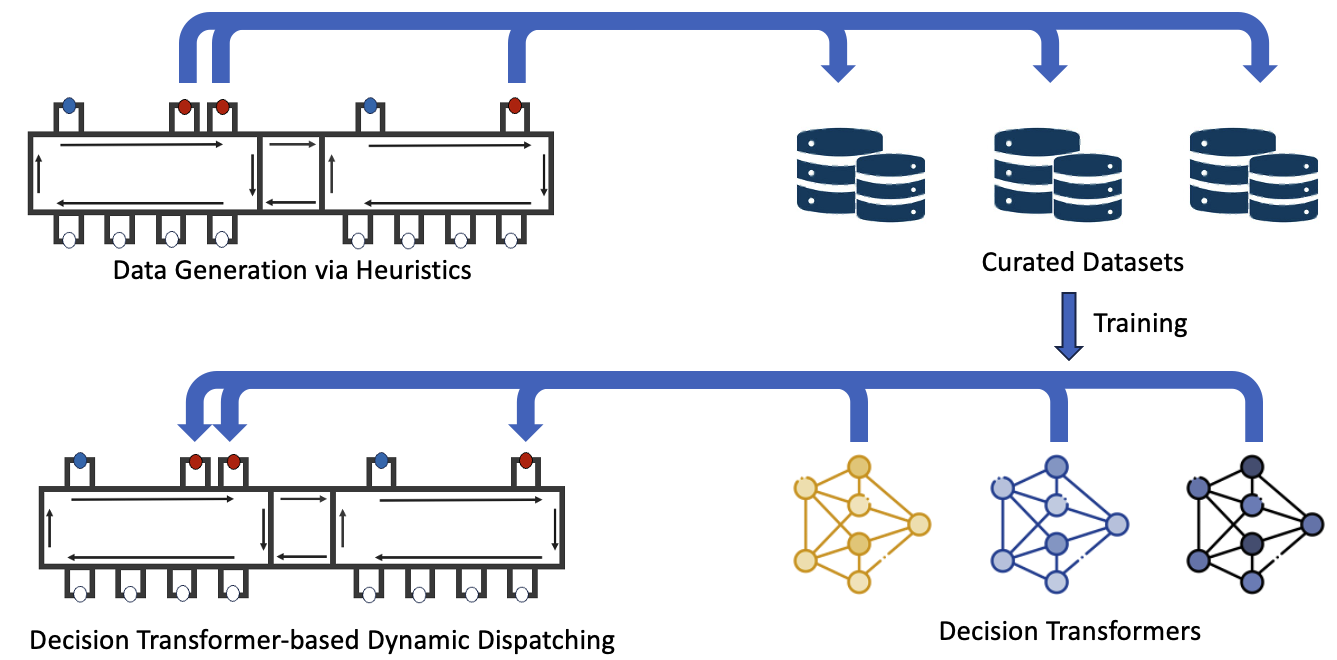}
    \caption{Overview of our proposed framework of using multiple Decision Transformers as dynamic dispatching policies in a multi-agent setting. }
    \label{fig:framework}
\end{figure*}

\section{Formulation and Methodology}

\subsection{Formulation}

This section presents the formulation that allows us to apply Decision Transformers as dynamic dispatching policies. Formally, the problem of dynamic dispatching in material handling systems can be defined as selecting a sequence of decisions at each incoming point such that the maximum throughput of the system is maximized. This, in turn, could be abstracted into an RL formulation, where an RL-based policy observes the state of the system and selects an action, representing the dispatching decision, to maximize a specific reward, in this case, the total throughput. With the dynamic dispatching problem cast as an RL problem, we can then follow the conventional routine of collecting/generating a static dataset consisting of tuples of \textit{(state, action, reward, done)} and train off-the-shelf offline RL algorithms on it. In this study, we focus on Decision Transformers due to their powerful capabilities in attending to sequential data and relative ease of implementation.

Another design choice we made in this formulation is the choice of representation. As there are multiple incoming points where dispatching occurs, we face a choice of representing the dispatching policy using a single centralized Decision Transformer versus multiple decentralized Decision Transformers. Since dispatching at the incoming points occurs asynchronously, we hypothesize that representing the dispatching policies in a multi-agent paradigm will be more beneficial for several reasons. The first reason is that if we use a centralized Decision Transformer, intuitively, the model would have to learn not just to model the joint distributions of states and actions but also to model the probability of an event occurring at a certain incoming point, which is significantly more challenging. Additionally, a centralized Decision Transformer limits the approach's scalability, as adding incoming points to the system's layout would require retraining the model to account for the change in action space. In contrast, using multiple Decision Transformers simplifies the modeling task and potentially allows us to re-use the same trained model for new incoming points since conveyor layouts are often replicated in material handling systems. Nevertheless, using a multi-agent paradigm also has its drawbacks since that inherently requires training multiple models and potentially introduces the problem of coordination between different models. Figure~\ref{fig:framework} provides an overview of the framework in this paper, which represents each incoming point's dynamic dispatching policy using a separate Decision Transformer in a multi-agent setting.

%
\subsection{Environment and Heuristics Details}

Due to confidentiality restrictions preventing the use of actual enterprise data for publication, we developed an in-house high-fidelity simulator that utilizes data distributions obtained from actual enterprise data to replicate the actual material handling system operations and generate the data for training Decision Transformers. We implemented multiple heuristics that have been tested on the actual system in the simulator and verified that the simulator's specifications and properties, such as the processing times and throughput match the actual system. After validating that the simulator is calibrated to the actual system faithfully, we used the simulator to generate data for training the Decision Transformers. Although we used simulated data in this work, we emphasize that the framework remains directly applicable to actual data. To collect the simulation data in a usable form, we define the simulator's state as all the sensor information that is available in real-time and used by the existing heuristics: \textbf{1)} Number of pallets heading to each storage point, \textbf{2)} Number of pallets at conveyor's junction going into each downstream direction, \textbf{3)} Inventory level at each storage, which leads to a 44-dimensional state vector. The action space is defined as the dispatching decisions and the reward is the total throughput of the system. The full details of the actual system's layout are shown in Table~\ref{tab:spec}

\begin{table}[htbp]
\caption{Material Handling System Specifications}
\begin{center}
\begin{tabular}{l|c}
\hline
\textbf{Specification} & \textbf{Value} \\
\hline
Number of Loops & 3  \\
Number of Incoming Points & 4  \\
Number of Storage Points & 20  \\
Number of Outgoing Points & 6  \\
Number of Junction Points & 4  \\
Incoming Points Processing Time & 5 sec  \\
Storage Points Processing Time & 10 sec  \\
Outgoing Points Processing Time & 6 sec  \\
Junction Points Processing Time & 0.5 sec  \\
Buffer size for Incoming Points & 4\\
Buffer size for Storage Points & 8\\
Buffer size for Outgoing Points & 10\\
Number of Pallets & 500 \\
Simulation resolution & 0.1 sec/step \\
\hline
\end{tabular}
\label{tab:spec}
\end{center}
\end{table}

Following the convention of the baselines presented in ~\cite{lee2024multi}, we implemented four heuristics with different skill levels, denoted as  'Low', 'Medium', 'High', based on the observed performance, and 'Random'.  We first define several auxiliary variables and functions that are employed across these heuristics. Each storage point in the system is associated with a specific "loop" in the conveyor system. For any incoming-storage point pair, we categorize the storage points as either being in the same loop as the incoming point (\(\mathbb{S}_{\text{same}}\)) or in different loops (\(\mathbb{S}_{\text{other}}\)). We also define \(\mathbb{S}_{\text{all}} = \mathbb{S}_{\text{same}} \cup \mathbb{S}_{\text{other}}\) as the set of all storage points. Let \( \text{In}(s) \) denote the number of incoming pallets and \( \text{Out}(s) \) denote the number of outgoing pallets at storage point \( s \). Additionally, we define \( X_{\text{same}} \) and \( X_{\text{other}} \) as the number of pallets assigned to storage points in the same loop and other loops, respectively, for a given incoming point. In the second heuristic, we introduce a cost function \( \text{cost}(L_i) \), which is defined as:

\[
\text{cost}(L_{i}) = \frac{X_{\text{same}} - X_{\text{min}}}{X_{\text{max}} - X_{\text{min}}} + C_{L_{j}}
\]

where $L_{i}$ denotes a specific loop $i$, \( X_{\text{max}} \) and \( X_{\text{min}} \) represent the maximum and minimum number of pallets across all loops, and \( C_{L_{j}} \) is a hyperparameter that reflects the distance cost between loops \( i \) and \( j \).

The three heuristics are summarized as follows: The first heuristic randomly assigns pallets to storage points within the same loop (\(\mathbb{S}_{\text{same}}\)), based on proximity. The second heuristic selects storage points based on a combination of buffer occupancy, distance, and pallet flow (inbound/outbound), using the cost function to prioritize loops. Intuitively, it selects storage points by filtering based on expected buffer occupancy, the cost function, and then choosing the storage point with the fewest incoming pallets. The third heuristic extends the second heuristic by adding logic to handle loop congestion, filtering by pallet distribution across loops and expected buffer occupancy before finally selecting a storage point based on the smallest difference between outgoing and incoming pallets. The algorithms corresponding to these heuristics, referred to as Low, Medium, and High, are detailed in Algorithms~\ref{alg:h1}, \ref{alg:h2}, and \ref{alg:h3}, respectively. These heuristics were developed through expert's knowledge and fine-tuned using simulation models to optimize performance. Finally, we also implemented a 'Random' heuristic, which is a simple baseline that dispatches the pallets to random storage points.

For dispatching decisions at junctions, a heuristic directs empty pallets towards conveyor loops with the fewest pallets.

\begin{algorithm}
\caption{Heuristic 1 (Low)}\label{alg:h1}
\SetAlgoLined
\KwIn{\texttt{Environment}, Number of Episodes $N$, Episode Horizon $T$}
\KwResult{Action decisions $a_t$ at each time step}
\For{Episode = 1 to $N$}{
    \For{$t = 1$ to $T$}{
        Observe state $s_t$ and event indicator $I_t$ from \texttt{Environment}\;
        \eIf{$I_t$ is True}{
            $\mathbb{S} \gets \mathbb{S}_{\text{same loop}}$ \tcp{Get set of storage points within the same loop}
            $a_t \sim \mathbb{S}$ \tcp{Sample and dispatch to random storage point}
        }{
            Skip $a_t$ \tcp{Non-event transition}
        }
    }
}
\end{algorithm}

\begin{algorithm}
\caption{Heuristic 2 (Medium)}\label{alg:h2}
\SetAlgoLined
\KwIn{\texttt{Environment}, Number of Episodes $N$, Episode Horizon $T$}
\KwIn{Parameter $C_1$}
\KwResult{Action decisions $a_t$ at each time step}
\For{Episode = 1 to $N$}{
    \For{$t = 1$ to $T$}{
        Observe state $s_t$ and event indicator $I_t$ from \texttt{Environment}\;
        \eIf{$I_t$ is True}{
            $\mathbb{S} \gets \mathbb{S}_{\text{all}}$ \tcp{Get set of all storage points}
            $\mathbb{S} \gets \{s \in \mathbb{S} \mid \text{In}(s) \leq C_1\}$ \tcp{Get storage points with fewer than $C_1$ incoming pallets}
            $\mathbb{S} \gets \text{minCost}(\mathbb{S})$ \tcp{Get storage points in loop with minimum cost}
            \eIf{$|\mathbb{S}| = 1$}{
                $a_t \gets s \in \mathbb{S}$ \tcp{If only one storage point in set, select it}
            }{
                $a_t \gets \arg\min_{s \in \mathbb{S}} (\text{In}(s))$ \tcp{Select storage with smallest number of incoming pallets}
            }
        }{
            Skip $a_t$ \tcp{Non-event transition}
        }
    }
}
\end{algorithm}

\begin{algorithm}
\caption{Heuristic 3 (High)}\label{alg:h3}
\SetAlgoLined
\KwIn{\texttt{Environment}, Number of Episodes $N$, Episode Horizon $T$}
\KwIn{Parameters $C_1$, $C_2$, $C_3$}
\KwResult{Action decisions $a_t$ at each time step}
\For{Episode = 1 to $N$}{
    \For{$t = 1$ to $T$}{
        Observe state $s_t$ and event indicator $I_t$ from \texttt{Environment}\;
        \eIf{$I_t$ is True}{
            $X_{\text{same}} \gets$ Number of pallets assigned to storage points in the same loop\;
            $X_{\text{other}} \gets$ Number of pallets assigned to storage points in other loops\;
            \eIf{$X_{\text{same}} < C_1$ \textbf{and} $X_{\text{other}} < C_2$}{
                $\mathbb{S} \gets \mathbb{S}_{\text{all}}$ \tcp{Get set of all storage points}
            }{
                \eIf{$X_{\text{same}} < C_1$ \textbf{and} $X_{\text{other}} > C_2$}{
                    $\mathbb{S} \gets \mathbb{S}_{\text{same}}$ \tcp{Get set of storage points in the same loop}
                }{
                    \eIf{$X_{\text{same}} > C_1$ \textbf{and} $X_{\text{other}} < C_2$}{
                        $\mathbb{S} \gets \mathbb{S}_{\text{other}}$ \tcp{Get set of storage points in other loops}
                    }{
                        $\mathbb{S} \gets \mathbb{S}_{\text{all}}$ \tcp{Get set of all storage points}
                    }
                }
            }
            $\mathbb{S} \gets \{s \in \mathbb{S} \mid \text{In}(s) \leq C_3\}$ \tcp{Get storage points with fewer than $C_3$ incoming pallets}
            \If{$\mathbb{S} \setminus \{S_{\text{others}}\} \neq \varnothing$}{
                $\mathbb{S} \gets \mathbb{S} \setminus \{S_{\text{others}}\}$ \tcp{Remove storage points that belong to other loops}
            }
            \eIf{$|\mathbb{S}| = 1$}{
                $a_t \gets s \in \mathbb{S}$ \tcp{If only one storage point in the set, select it}
            }{
                $a_t \gets \arg\min_{s \in \mathbb{S}} (\text{Out}(s) - \text{In}(s))$ \tcp{Select storage point with minimum out-in difference}
            }
        }{
            Skip $a_t$ \tcp{Non-event transition}
        }
    }
}
\end{algorithm}

\subsection{Data Generation}

For each heuristic, we ran 4000 one-hour simulations, corresponding to 4000 episodes. A one-hour simulation time period was selected because it represents a time-frame with sufficient events occurring and is also a commonly used time-frame used for comparing throughput in the industry.  Specifically, at each incoming process point, we collected the data in the form of \textit{(state, action, reward, done)} whenever an event, i.e., a dispatching decision is needed, occurs at the specific point. Since the events occur asynchronously for each incoming point, the total number of events at each incoming point is different for each episode. Nevertheless, we observed that the number of events at each incoming point falls within the range of 500 to 550 events in each episode. In summary, we curated four different datasets corresponding to four different heuristics, each consisting of 4000 trajectories for each incoming point, yielding approximately a dataset of 2 million transitions to train each Decision Transformer for each dispatching point. All simulations were performed on a 24-core Intel(R) Core(TM) i9-10920X CPU.

\subsection{Training}

We trained a Decision Transformer for each dispatching point in the material handling system on each generated dataset and trained each model for 50 epochs based on the observed convergence and plateauing of the loss function. Since the dynamic dispatching in this problem necessitates discrete actions, we trained the models using a standard cross-entropy loss. Our implementation is based on a modification of the implementation by~\cite{dt_hf}. All training was performed on a single NVIDIA RTX 4090 GPU and the hyperparameters we used in our experiments are default implementation parameters and are shown in Table~\ref{tab:hyperparameters}

\begin{table}
\centering
\begin{tabular}{l|c}
\hline
\textbf{Hyper parameters} \\
\hline
Max past sequence length ($k$) & 20 \\
Optimizer & AdamW  \\
Batch size & 256 \\
Warm up ratio & 0.1 \\
Max. grad. norm & 0.25  \\
Learning rate & 1e-4 \\
Weight decay & 1e-4 \\
Number of train epochs & 50 \\
\hline
\end{tabular}
\caption{Decision Transformer hyperparameters}
\label{tab:hyperparameters}
\end{table}

\section{Results and Discussion}

\subsection{Results}

In this section, we present empirical results of our experiments and answer the questions we posed.

\begin{figure}[!h]
    \centering
    \begin{subfigure}[t]{0.48\textwidth} 
        \centering
        \includegraphics[width=\textwidth]{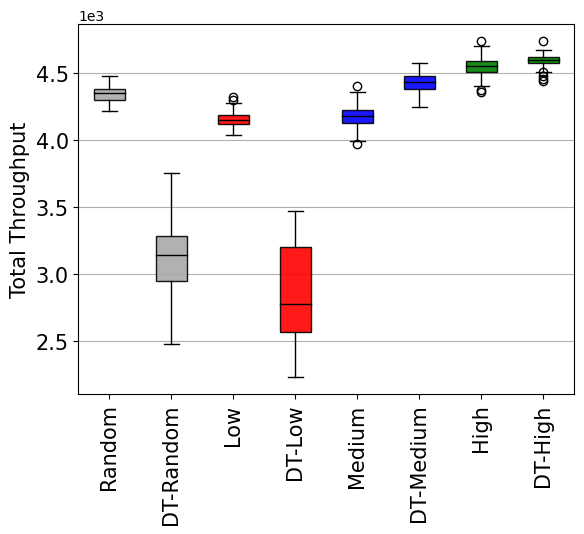}
        \caption{}
        \label{fig:main1}
    \end{subfigure}
    \hfill
    \begin{subfigure}[t]{0.48\textwidth} 
        \centering
        \includegraphics[width=\textwidth]{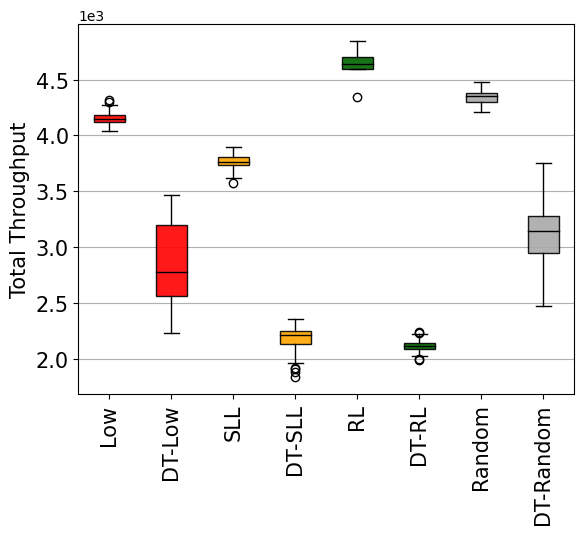}
        \caption{}
        \label{fig:main2}
    \end{subfigure}
    \caption{(a) Distribution of throughput between heuristics of different skill levels and Decision Transformers trained on the data generated by the heuristics. (b) Additional comparison of the throughput distributions of Decision Transformers trained on data generated by a deterministic low-skill heuristic (SLL) and on data generated by a stochastic high-skill online RL policy.}
    \label{fig:main}
\end{figure}

\begin{figure}[htbp]
    \centering
    \includegraphics[width=0.5\columnwidth]{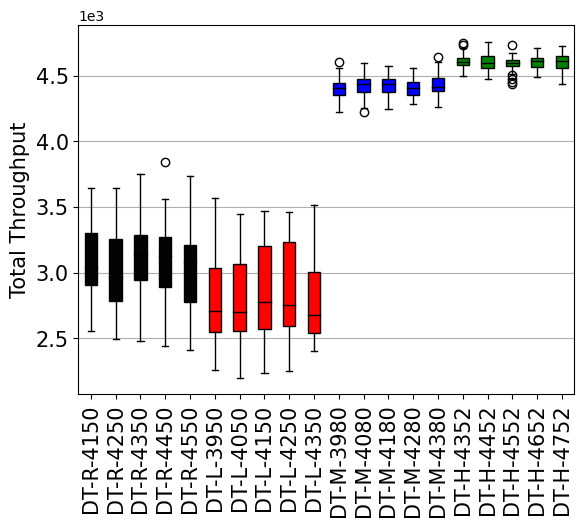}
    \caption{Distribution of throughput of Decision Transformers of different skill levels conditioned on different specified throughput. 'DT-R', 'DT-L', 'DT-M', 'DT-H' denote models that are trained on the 'Random', 'Low', 'Medium' and 'High' datasets respectively.}
    \label{fig:main_2}
\end{figure}

\textbf{Can Decision Transformers be applied to dynamic dispatching problems in a multi-agent asynchronous setting, and can they ``stitch” lower reward trajectories from potentially sub-optimal heuristics to achieve high rewards during testing?}

To answer the question above, we compare the total throughput of the Decision Transformers as dispatching policies with the corresponding heuristics' performance used to generate the training data across 50 episodes and 5 random seeds. During testing, we used the median throughput achieved by the heuristics as the desired return rather than using an arbitrary return. Figure~\ref{fig:main}(a), presents the results of the comparison. We observe that for experiments where the Decision Transformers were trained on data collected under the 'Medium' and 'High' heuristics, Decision Transformer policies outperform the heuristics' original performance. Conversely, for Decision Transformers that were trained using data generated from 'Random' and 'Low' heuristics, the Decision Transformers severely under-perform the heuristics.

We hypothesize that the Decision Transformers trained on the 'Low' and 'Random' data performed poorly due to two possible reasons: 1.) First, the 'Random' and 'Low' heuristics were both rules that contain an inherent aspect of randomness in the decision-making. In contrast, the 'Medium' and 'High' heuristics were fully deterministic rules. Hence, one possibility is that Decision Transformers do not perform well when trained on data generated with some aspect of inherent randomness. 2) Secondly, another possibility is there is a certain threshold of performance that needs to be exhibited by the trajectories of the heuristics for the Decision Transformer to stitch together to generate better trajectories. Since Decision Transformer is an offline method and is inherently data-dependent, this could also be a possibility that explains the poor performance of the Decision Transformer trained on the 'Low' heuristic's data. 

To verify our hypothesis, we conducted two additional experiments. We designed a fourth heuristic with a relatively low performance while keeping the logic fully deterministic. Specifically, we designed the heuristic to select storage points in the same loop as the incoming points and send the pallets to the storage points with the least number of pallets that are assigned to them. This is similar to the 'Low' heuristic, except that instead of assigning to a random storage point, we assign it to the least busy storage point. We denote this heuristic as 'SLL', representing ``Same Loop-Least". This experiment is to study the impact of data with low performance on the training of the Decision Transformer. Additionally, we also conducted another experiment where we trained an online RL algorithm in the simulator using a multi-agent PPO algorithm~\cite{yu2022surprising}. Subsequently, we use this online policy to generate a static dataset for training Decision Transformers following the same procedures above. The purpose of this experiment is to determine if randomness affect the performance of the trained Decision Transformer, since the online RL method is able to explore and collect more trajectories to achieve a higher performance, while the action generated by the RL algorithm is sampled from a learned distribution. We denote this method as 'RL'. We refer readers interested in the details of this approach to the following work~\cite{yu2022surprising}.

Figure~\ref{fig:main}(b), shows the results of the additional experiments, with the results of the 'Low' and 'Random' heuristics shown for additional reference. Comparing the 'SLL' with the 'Low' heuristic, we observe that the 'SLL' heuristic performance is slightly lower than the 'Low' heuristic. However, despite the heuristic being deterministic, we see that the Decision Transformer trained on the 'SLL' data (DT-SLL) still performs worse than the heuristic. 
This confirms the hypothesis that even in the absence of randomness, data with high performance trajectories is needed to train a Decision Transformer that outperforms the heuristic. On the other hand, we can see that an RL agent trained in an online fashion is able to achieve a high throughput value, similar to the performance of the 'High' heuristic. Nonetheless, despite the data generated by the RL policy having high performance, the Decision Transformer trained on those data also significantly under performs the heuristic's throughput value. This highlights another limitation that the data generated needs to be deterministic as well, confirming our earlier hypothesis.

Based on the the results presented, we anecdotally conclude that in this case, Decision Transformers can 'stitch' multiple reward trajectories to emit high reward trajectories, subject to the fact that the data these models are trained on were not generated with some aspect of inherent randomness and the reward of the trajectories have to be over a certain threshold, rather than trajectories with any arbitrary performance. Table~\ref{tab:main_results} tabulates the summary statistics of the box plots shown in Figure~\ref{fig:main}.

\begin{table}[h]
    \centering
    \begin{tabular}{l|c|c|c|c|c}
        \hline
        \textbf{Experiments} & \textbf{Min} &\textbf{1st Q}&\textbf{Median}&\textbf{3rd Q}&\textbf{Max}\\
        \hline
        Random & 4213 & 4301  & 4349 & 4380 & 4479 \\
        Low & 4040 & 4118 & 4150 & 4183 & 4317\\
        Medium & 3970 & 4128 & 4180 & 4221 & 4405 \\
        High & 4357 & 4505 & 4552 & 4589 & 4738 \\
        SSL & 3624 & 3731 & 3765 & 3805 & 3897 \\
        RL & 4340 & 4594 & 4642 & 4700 & 4845\\
        \hline 
        DT-Random & 2480 & 2946  & 3142 & 3283 & 3750\\
        DT-Low & 2233 & 2568 & 2777 & 3201 & 3471\\
        DT-Medium & 4246 & 4376 & 4432 & 4473 & 4571 \\
        DT-High & 4510 & 4574 & 4599 & 4620 & 4670 \\
        DT-SSL & 1965 & 2133 & 2214 & 2255 & 2355 \\
        DT-RL & 2027 & 2087 & 2121 & 2143 & 2225\\
        \hline
    \end{tabular}
    \caption{Numerical statistics of box-plots shown in results.}
    \label{tab:main_results}
\end{table}

Additionally, despite the Decision Transformers trained on data derived from deterministic high-performing heuristics outperforming their corresponding heuristics, we note that the achieved total throughput does not match the specified desired total throughput. To further investigate the behaviors of these Decision Transformers, we conducted five additional experiments by conditioning the models based on a range of values centered around the median throughput (specifically, median $\pm$ 100, 200). Figure~\ref{fig:main_2} shows the experiment results. Empirically, we observe that regardless of the types of heuristics used to generate the training data, the Decision Transformers showed no clear correlation between the specified desired rewards and actual rewards. 

\textbf{Are Decision Transformers affected by data with inherent stochasticity?}

\begin{figure}[!h]
    \centering
    \includegraphics[width=0.95\textwidth]{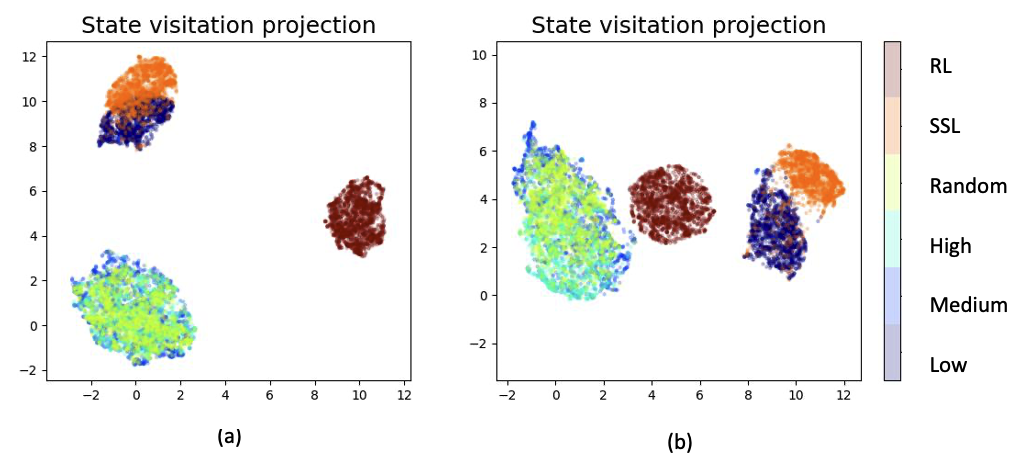}
    \caption{UMAP projection of the unique state visitations generated by the six heuristics of different skill level in (a) the $1^{st}$ and $2^{nd}$ dimension and (b) the $2^{nd}$ and $3^{rd}$ dimension.}
    \label{fig:umap}
\end{figure}

\begin{figure}[!h]
    \centering
    \begin{subfigure}[b]{0.48\textwidth} 
        \centering
        \includegraphics[width=\textwidth]{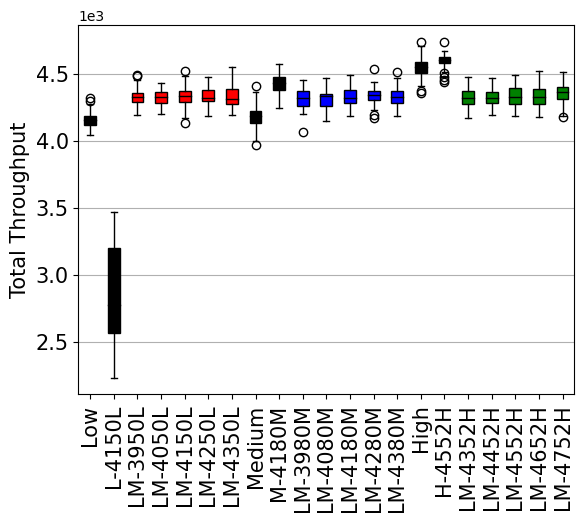}
        \caption{}
        \label{fig:sub1}
    \end{subfigure}
    \hfill
    \begin{subfigure}[b]{0.48\textwidth} 
        \centering
        \includegraphics[width=\textwidth]{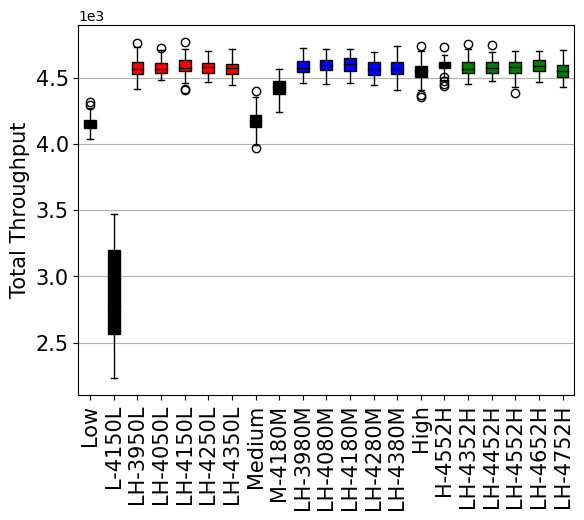}
        \caption{}
        \label{fig:sub2}
    \end{subfigure}

    \begin{subfigure}[b]{0.48\textwidth} 
        \centering
        \includegraphics[width=\textwidth]{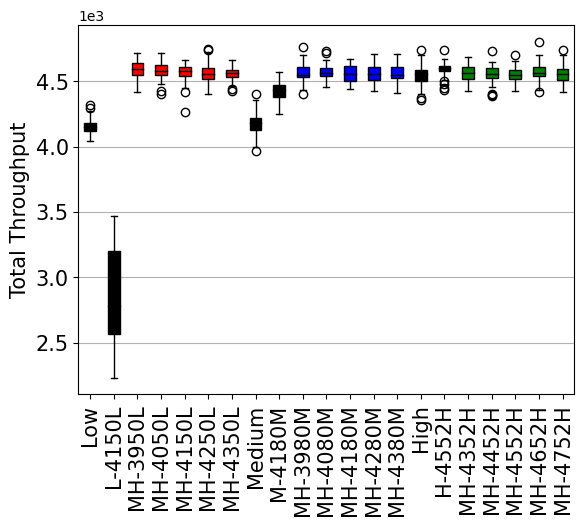}
        \caption{}
        \label{fig:sub3}
    \end{subfigure}
    \hfill
    \begin{subfigure}[b]{0.48\textwidth} 
        \centering
        \includegraphics[width=\textwidth]{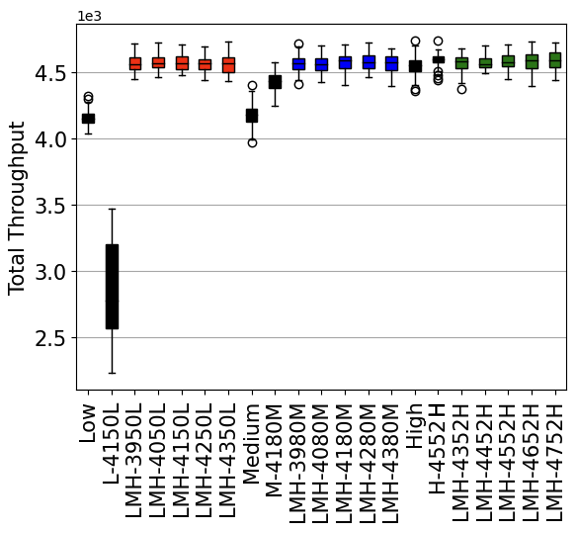}
        \caption{}
        \label{fig:sub4}
    \end{subfigure}
    \caption{Comparison of throughput for heuristics, Decision Transformers trained on a single dataset and Decision Transformers trained on datasets generated from multiple heuristics, conditioned on a range of desired rewards. The postfix 'L','M','H' denotes the specific heuristic from which the range of desired rewards was determined. (a) 'LM' denotes dataset with trajectories from the 'Low, Medium' data, (b) 'LH' denotes dataset with trajectories from the 'Low, High' data, (c) 'MH' denotes dataset with trajectories from the 'Medium, High' data, (d) 'LMH' denotes dataset with trajectories from the 'Low, Medium, High' data.}
    \label{fig:spectrum}
\end{figure}

As alluded to in the previous paragraph, we have demonstrated that Decision Transformers are affected by stochasticity in terms of action data generated by the heuristic. However, we also highlight that the state transitions generated in our dataset are also stochastic, due to randomness in terms of the goods being retrieved from the storage points. As such, based on our observations in Figure~\ref{fig:main}, we can only anecdotally confirm that a vanilla implementation of Decision Transformers is less affected by state stochasticity than action stochasticity. Nevertheless,~\cite{paster2022you} has also demonstrated that stochasticity in states could contribute to the lack of alignment between the specified and achieved rewards and that a better reward could be achieved by clustering the trajectories and conditioning the models on the average cluster rewards during training. We intend to investigate this direction in future works. 

\textbf{How does data quality affect the performance of Decision Transformer?}

In this section, we conduct a study into how the quality of data affects the final performance of Decision Transformers. As a first step, we conducted a dimensionality reduction of states in our datasets to visualize the unique state visitations in our datasets. To achieve that, we removed all duplicates of states in the collected datasets and projected them into a 3-dimensional space using UMAP~\cite{mcinnes2018umap}. Figure~\ref{fig:umap}(a) and (b) show the projection of high-dimensional states into the $1^{st}$-$2^{nd}$ and $2^{nd}$-$3^{rd}$ dimensions, respectively, with the colors corresponding to data generated with the six different heuristics. We observed that the state visitations of the data generated by the 'Random,' 'Medium,' and 'High' form a relatively tight cluster in the first two dimensions. In contrast, the 'Low' and 'SLL' data forms two relatively separate clusters that is distant from the rest of the data. Finally, the data generated by the high-performing online RL approach also forms a fourth distinct cluster of state visitations. A similar observation can be also be made in the $2^{nd}$-$3^{rd}$ dimension of UMAP projection as seen in Figure~\ref{fig:umap}(b).
The main takeaway of these figures is that different heuristics emit state visitations that cover different regions of the state space and training Decision Transformers on data generated by a single heuristics may be sub-optimal since the coverage of states visited due to any single, manually crafted heuristics could potentially be limited.

Next, we study the effect of combining datasets with different skill levels on the performance of the Decision Transformers. Specifically, we curated multiple additional datasets with 12,000 trajectories each, denoted as 'LMH,' 'LH,' 'LM,' and 'MH.' The 'LMH' data is made up of 4000 trajectories sampled from the 'Low,' 'Medium' and 'High' datasets each, while the 'LH,' 'LM,' and 'MH' data is made up of 6000 trajectories sampled from the 'Low-High,' 'Low-Medium' and 'Medium-High' data respectively. We re-trained the Decision Transformers based on these data and conditioned the models based on a range of throughput values achieved by the 'Low,' 'Medium,' and 'High' heuristics, respectively. Figure~\ref{fig:spectrum} presents the results of the experiments. One apparent observation is that the inclusion of data generated by deterministic, high performing heuristics significantly outweighs the effect of data generated by heuristics with randomness or heuristics with low performance. This can be seen in Figures~\ref{fig:spectrum}(a),(b), and (d), where the models trained with a combination of data sampled from the 'Medium' or 'High' and 'Low' datasets significantly improve over the models trained on just the 'Low' dataset. We also observe that the Decision Transformers tend to be biased towards the performance of the 'High' dataset. For example, in Figures~\ref{fig:spectrum}(b) and (d), the Decision Transformers still fail to achieve the specified rewards belonging to the region of the 'Low' heuristics despite including an equal amount of data from the 'Low' dataset into the combined datasets. Furthermore, from Figure~\ref{fig:spectrum}(b), we observe that the Decision Transformer also failed to achieve the specified rewards of a 'Medium' heuristic, despite having trained on 'Low' and 'High' data and having seen a wider coverage of state visitations as shown in Figure~\ref{fig:umap}(a). In summary, we conclude that in our application, Decision Transformers tend to be biased towards high-performing data, generally fail to interpolate between the datasets to achieve unseen target returns, and their performance does not seem to improve with a wider quality of data. However, this might be specific to our application and may be attributed to one or more features, such as being in a multi-agent setting and the presence of stochasticity in the state and action data. 

\subsection{Discussions}
In this section, we discuss the implications of this work from an industry point of view. Historical data owned by enterprises in the industrial sector holds substantial potential for training Decision Transformers for decision-making systems in an offline manner. Our study highlights that while Decision Transformers hold promise, their effectiveness is significantly influenced by the randomness in the underlying dataset and sub-optimal data can severely limit their effectiveness. To this end, we posit that filtering/ranking techniques and statistical methods could potentially be applied to the dataset to obtain a subset of higher-performing data and to detect the presence of randomness in the absence of domain experts knowledge of the enterprise data. This underscores the importance for industries to not only gather large volumes of data but also serves as a guideline to determine when training Decision Transformers will result in effective dynamic dispatching strategies.

One of the main benefits of Decision Transformers or any offline RL approach in general for decision-making problems is the ability to circumvent the requirement of developing a high-fidelity simulator or training the agent on an actual system. However, to validate the effectiveness of these methods in actual industrial applications, a simulator is still often required to test and refine the policies as it is often too costly and risky to directly deploy a dispatching policy on the system. Despite this, enterprises with extensive historical data may still find this method beneficial. From an implementation perspective, it is simpler to leverage historical data to train Decision Transformers to achieve marginal improvements due to less hyper-parameter tuning required, provided that the data meets the required conditions, as compared to an online RL approach. Nonetheless, deployment of Decision Transformer-based dispatching policies on to an actual system is also susceptible to the challenges that are common to most data-driven machine learning methods, such as integration with existing infrastructure, distribution drifts and the occurrence of edge cases. While these challenges are non-trivial, we believe that they're not specific to Decision Transformers or dynamic dispatching problems and could potentially be overcome by solutions such as a meticulous system integration plans, real-time data monitoring platforms and careful design of rules for exception handling, e.g., data transformation pipelines that ensure input and output data lies within an acceptable range. 

Our findings highlighted other limitations of the current implementation of Decision Transformers, particularly the lack of a strong correlation between the specified target return and the actual return achieved. This points to the need for further experimentation to isolate the cause of these issues as well as a study into more sophisticated versions of Decision Transformers, such as those using reward prediction mechanisms, to address these shortcomings. We believe that future studies focusing on how existing data can be better utilized or augmented with domain knowledge to enhance the capabilities of Decision Transformers, would be beneficial to improve their application in industrial automation settings. 

\section{Conclusion}

We show that Decision Transformers have the potential to be deployed as dynamic dispatching policies in material handling systems under the assumption that existing enterprise data can be transformed into the required format and that the underlying method generating the data is not inherently random and has a good initial performance. Due to the relatively simple implementation of Decision Transformers, this has the potential to democratize the deployment of these models in actual industrial systems to improve business KPIs such as system throughput without the need to train online RL-based methods. Future works will focus on complementary improvements such as methods that could better incorporate probabilistic data into the training process, better communication strategies between the multiple agents, perhaps through a centralized critic and also hyperparameter tuning to improve the existing implementation of Decision Transformers. Additionally, benchmarking the performance of Decision Transformers against other state-of-the-art offline RL methods and testing the generalization capability of multi-agent Decision Transformers to other dynamic dispatching systems are also important avenues of study. 

\bibliographystyle{unsrtnat}
\bibliography{references}  

\end{document}